\documentclass[conference]{IEEEtran}
\usepackage{blindtext}
\usepackage{eso-pic}
\IEEEoverridecommandlockouts
\usepackage{cite}
\usepackage{amsmath,amssymb,amsfonts}
\usepackage{algorithmic}
\usepackage{graphicx}
\usepackage{textcomp}
\usepackage{xcolor}

\usepackage[export]{adjustbox}
\usepackage[utf8]{inputenc} 
\usepackage[T1]{fontenc}    
\usepackage{hyperref}       
\usepackage{url}            
\usepackage{booktabs}       
\usepackage{amsfonts}       
\usepackage{nicefrac}       
\usepackage{microtype}      
\usepackage{xcolor}         
\usepackage{acro}
\newtheorem{task}{Task}
\usepackage{amsmath,amssymb,multicol,latexsym}
\usepackage{breqn}
\usepackage{pgfplots,pgfplotstable}
\pgfplotsset{compat=1.16}
\usepackage{tikz}
\usepackage{cite}
\usetikzlibrary{shapes,
 	automata,
 	arrows,
 	chains,
 	matrix,
 	backgrounds,
 	fit,
 	patterns,
 	decorations.markings,
 	svg.path,
 	shapes.multipart,
 	shapes.symbols,
 	external,
 	shadows,
 	positioning,
 	calc,
 	decorations,
 	scopes,
 	patterns,
 	patterns.meta,
 	bending}
 \usetikzlibrary{external}

\usepackage{hyperref}
\usepackage{cleveref}
\usepackage{flushend}
\usepackage{comment}

\DeclareMathAlphabet{\mathcal}{OMS}{cmsy}{m}{n}

\usepackage{url}
\usepackage{graphicx}
\usepackage{svg}
\usepackage{numprint,fullpage}
\usepackage{booktabs}
\usepackage{multirow}

\usepackage{adjustbox}

\Crefname{equation}{Eq.}{Eqs.}
\Crefname{figure}{Fig.}{Figs.}
\Crefname{tabular}{Tab.}{Tabs.}
\usepackage{tabu}
\usepackage{anysize} 
\graphicspath{{figures/}}
\def\BibTeX{{\rm B\kern-.05em{\sc i\kern-.025em b}\kern-.08em
    T\kern-.1667em\lower.7ex\hbox{E}\kern-.125emX}}

\DeclareAcronym{IUTQ}{
    short = TQ,
    long = Inverse Universal Traffic Quality
}

\DeclareAcronym{TTC}{
    short = TTC,
    long = time-to-collision
}

\DeclareAcronym{WTTC}{
    short = WTTC,
    long = worst time-to-collision
}

\DeclareAcronym{PTTC}{
    short = PTTC,
    long = predictive time-to-collision
}

\DeclareAcronym{PET}{
    short = PET,
    long = post-encroachment time
}

\DeclareAcronym{ET}{
    short = ET,
    long = encroachment time
}

\DeclareAcronym{GT}{
    short = GT,
    long = gap time
}

\DeclareAcronym{MCC}{
    short = MCC,
    long = Matthews correlation coefficient
}

\DeclareAcronym{F1S}{
    short = F1S,
    long = $F_1$-score
}

\DeclareAcronym{CoK}{
    short = CoK,
    long = Cohen's kappa
}

\graphicspath{{img/}}
\ifCLASSINFOpdf
\else
\fi
\usepackage{url}

\begin{document}
%
\title{Inverse Universal Traffic Quality - a Criticality Metric for Crowded Urban Traffic Scenes}


\author{\IEEEauthorblockN{Barbara Schütt\IEEEauthorrefmark{1}\IEEEauthorrefmark{2}, Maximilian Zipfl\IEEEauthorrefmark{1}\IEEEauthorrefmark{2}, J.~Marius~Zöllner\IEEEauthorrefmark{1}\IEEEauthorrefmark{2} and Eric Sax\IEEEauthorrefmark{1}\IEEEauthorrefmark{2}}
\IEEEauthorblockA{\IEEEauthorrefmark{1}FZI Research Center for Information Technology, Karlsruhe, Germany\\
Email: {schuett, zipfl, zoellner, sax}@fzi.de}
\IEEEauthorblockA{\IEEEauthorrefmark{2}Karlsruhe Institute of Technology, Karlsruhe, Germany}}
\maketitle

\begin{abstract}
An essential requirement for scenario-based testing the identification of critical scenes and their associated scenarios.
However, critical scenes, such as collisions, occur comparatively rarely.
Accordingly, large amounts of data must be examined.
A further issue is that recorded real-world traffic often consists of scenes with a high number of vehicles, and it can be challenging to determine which are the most critical vehicles regarding the safety of an ego vehicle.
Therefore, we present the inverse universal traffic quality, a criticality metric for urban traffic independent of predefined adversary vehicles and vehicle constellations such as intersection trajectories or car-following scenarios.
Our metric is universally applicable for different urban traffic situations, e.g., intersections or roundabouts, and can be adjusted to certain situations if needed.
 Additionally, in this paper, we evaluate the proposed metric and compares its result to other well-known criticality metrics  of this field, such as time-to-collision or post-encroachment time.
\end{abstract}

\IEEEpeerreviewmaketitle

\section{Introduction}

Scenario-based testing has become increasingly crucial for highly automated driving, particularly in the European Union. 
On August 5, 2022, the implementation regulation \cite{EUtypeapprovannex} regarding EU Regulation No. 2019/2144 \cite{EUtypeapprov} went into effect and became mandatory in all EU member states. 
This regulation outlines a set of methods for assessing the overall compliance of an automated driving system.
It specifies that scenario-based testing with a minimum set of traffic scenarios that are relevant to the system's operating design domain should be used. 


One problem with scenario-based testing for automated and autonomous vehicles is using metrics to evaluate scenarios and traffic scenes to determine if a scene contains a serious conflict. 
Many criticality metrics focus on a specific feature that may not apply to all scenes or require specific conditions.
Furthermore, it is common to consider individual entities in a scene separately. 
These problems can make it challenging to accurately and comprehensively assess the performance of an automated or autonomous vehicle and ensure its safety on public roads.
In a previous paper \cite{schutt2022fingerprint}, we addressed this problem and examined the criticality of different scenes using several metrics. 
There, a multidimensional scoring model that can assess scenes based on standard metrics regardless of the scene type was introduced.
The basic assumptions and ideas for the \acf{IUTQ} metric were already presented in this context. 
The current work presents a comprehensive metric description, and an in-depth analysis based on real-world data is performed.
The main features of the metric proposed in this work are as follows:
\begin{itemize}
    \item[(i)] independent of vehicle constellation (e.g., intersection or car-following) and usable when more than two parties are involved,
    \item[(ii)] holistic approach for evaluation of a scene from the perspective of an ego vehicle
    \item[(iii)] reusability to address different research questions and requirements due to agnostic design and high parameterization capability.
\end{itemize}
\begin{figure}[t!]
    \centering
       \def\svgwidth{0.95\columnwidth}
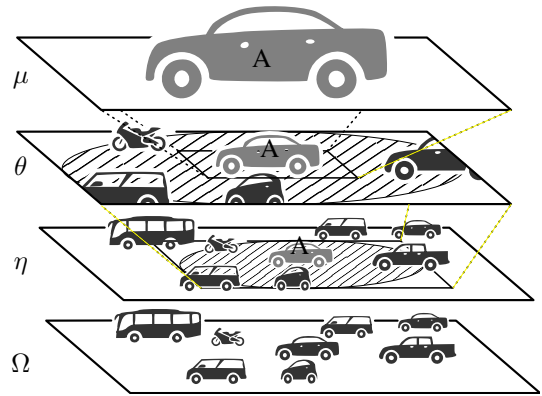
\caption{Overview of the different areas of interest of the Universal Inverse Traffic Quality on an exemplary traffic scene.}
\label{fig:overview_scheme}
\end{figure}
\Cref{sec:sota} gives a short introduction to the definitions of essential terms and background that serves as the base for the proposed metric. 
\Cref{sec:tq_metric} describes the metric and its sub-metrics, shown in a conceptional overview in \Cref{fig:overview_scheme}.
After that, \Cref{sec:evaluation} presents an evaluation and compares its performance to several established metrics.
\Cref{sec:conclusion} summarizes the most important findings of this work.
The implementation of this metric is added as a separate module in the criticality framework presented in earlier works \cite{schutt2022fingerprint}.
The open source code for the framework and implementation of this metric can be found on GitHub\footnote{\url{https://github.com/fzi-forschungszentrum-informatik/scene-fingerprint}}.
\section{Theoretical Background\\ \& State of the Art}
\label{sec:sota}
\subsection{Definitions}
Although there are several different definitions for the terms scene and scenario, this paper will follow the definitions of Ulbrich \textit{et al.} \cite{ulbrich_defining_2015}.
This means that a scene is described from all static and dynamic elements and the associated properties at a given point in time. 
Consequently, a scenario is a temporal sequence of successive scenes.
Since we objectively want to evaluate a scene and with that the quality of a scenario, the proposed metric lies in the category of scenario quality \cite{schutt2022taxonomy}.

Criticality is defined by Neurohr \textit{et al.} \cite{neurohr2021criticality} as the combined risk of the involved actors toward a crash or harmful events in case the traffic
situation is continued.
Further, criticality metrics are used to quantify the criticality in a given scene, situation, or scenario.
Junietz \cite{junietz2019microscopic} formulated a set of requirements for criticality metrics:
\begin{itemize}
    \item[R1] give a macroscopic risk statement that can be used in a safety argumentation for a concrete system, 
    \item[R2] identify critical scenarios within a dataset.
\end{itemize}
A list of known and established criticality metrics was compiled by Westhofen \textit{et al.} \cite{westhofen_criticality_2022}.
There, all metrics are briefly introduced and put into a common context.

\subsection{Highway Traffic Quality}
Hallerbach \textit{et al.} \cite{hallerbach_simulation-based_2018} propose a method to evaluate the criticality of the traffic on a highway section. 
The focus is not solely on the ego vehicle and one other traffic participant but also on different domains of interest, including larger road sections, and traffic around the ego vehicle.
The calculation of traffic quality is divided into four parts with differently weighted coefficients, which sum up to a total criticality score.
The first part, the \textit{macroscopic} metric, calculates the traffic density with the help of the traffic flow rate and the average travel velocity.
Additionally, the \textit{microscopic} metric considers the velocity deviation and the average velocity on a highway section around the ego vehicle.
The third part of the traffic quality term, the \textit{nanoscopic} metric, is centered on close-range interactions inside a circle around the ego vehicle with a smaller radius. 
Its calculation is based on velocity deviation and the mean value within it.
The last part is the \textit{individual} metric and concentrates on the ego vehicle's mean velocity and standard deviation of the acceleration.
Finally, a supervised training algorithm finds optimal weights as coefficients for all four terms, a procedure that has to be repeated for different scenes.
\section{Inverse Universal Traffic Quality}
\label{sec:tq_metric}
The \acl{IUTQ} is based on the traffic quality for highway scenarios as proposed by Hallerbach \textit{et al.} \cite{hallerbach_simulation-based_2018}.
A traffic scene is evaluated regarding different aspects of spatial and vehicle features.
At the focus of the metric is a traffic participant, the ego vehicle, to which all measured and calculated values refer.
A main difference to other metrics, e.g., \ac{TTC} or \ac{PET}, is that the whole scene is evaluated instead of only two traffic participants.
This approach makes \acl{IUTQ} better suitable for getting an overall impression of a crowded traffic scene.
However, instead of a final score calculated by scene-optimized weights, we offer a universal summary approach for the \acl{IUTQ}, which will be explained further in \Cref{sec:end_score}.

The following four sub-metrics have already been presented more superficially in our previous work \cite{schutt2022fingerprint}, although under a different naming.
Moreover, we use the term inverse since measured values increase with the growing criticality of a scene and, therefore, its quality is decreasing.
Additionally, information from previous time frames is avoided as much as possible to simplify computation.

 \subsection{Macroscopic Inverse Traffic Quality}
 The macroscopic traffic quality computes the coefficient of variation with respect to the velocities of all vehicles in a scene, i.e., the higher the macroscopic value, the more unsteady all traffic participants move.
 Uniform traffic is considered uncritical, regardless of the average velocity of all participants.
 The macroscopic \acl{IUTQ} for a complete scene is calculated with
 
 \begin{equation}
     \mathrm{TQ}_{\Omega} = \frac{\sigma_{sc}}{\bar{\nu}_{sc}}
 \end{equation}
 where $\sigma_{sc}$ describes the standard deviation of the velocities and $\bar{\nu}_{sc}$ the mean velocity of all vehicles $V_t$ within one scene at time $t$.
 The macroscopic value is the same for all traffic participants within one scene.
Typical situations with high macroscopic values are congestion situations, e.g., merging lanes or traffic lights, where most cars are standing still, and only a few are moving; a roundabout with vehicles waiting for a moment to enter; a high number of parking cars within a map.
The macroscopic sub-metric does not consider the spatial traffic density, e.g., an almost empty map with few vehicles with different velocities can also have high macroscopic values.

 \subsection{Metascopic Inverse Traffic Quality}
 The metascopic traffic quality is the ratio between all vehicles on the scene and the ones within the estimated braking distance of an ego vehicle $A$:
 
  \begin{equation}
     \mathrm{TQ}_{\eta}(A) = \frac{V_t^{br}(A)}
     {\left|V_t\right|}
 \end{equation}
where $\left|V_t\right|$ represents the set of all vehicles found on the scene at time $t$.
$V_t^{br}(A) \subseteq V_t $ where $V^{br}(A)$ denotes all vehicles which are within the breaking distance of $A$ at the time $t$.
The metascopic value always lies between $0$ and $1$ and rises with the ego vehicle's velocity until the braking distance includes the complete map of a scene.
Typical situations for a high metascopic value are fast movements across a map or leaving a scene with high acceleration.

\subsection{Mesoscopic Inverse Traffic Quality}
The mesoscopic traffic quality is similar to the macroscopic traffic quality. 
However, it only considers the coefficient of variation within the braking distance of an ego vehicle:

\begin{equation}
     \mathrm{TQ}_{\theta}(A) = \frac{\sigma_{br}}{\bar{\nu}_{br}}
 \end{equation}
where $\sigma_{br}$ describes the standard deviation of the velocities and $\bar{\nu}_{br}$ the mean velocity of all vehicles within the braking distance of the ego vehicle $V^{br}(A)$.
Typical situations for high mesoscopic values are where the ego vehicle passes or approaches a group of slower moving vehicles.
 
\subsection{Microscopic Inverse Traffic Quality}
The microscopic traffic quality is the only sub-metric that uses data from previous time frames.
Further, it only considers the ego velocity and acceleration development over the last few seconds.
It consists of two ratios combined:
  \begin{equation}
     \mathrm{TQ}_{\mu}(A) = \frac{\frac{\bar{a}_{A}}{a_{ref}} + \frac{\bar{\nu}_{A}}{\nu_{ref}}}{2}
 \end{equation}
where $\bar{a}_{A}$ describes the mean acceleration and $\bar{\nu}_{A}$ the mean velocity of the ego vehicle over a certain time in the past.
Values $a_{ref}$ and $\nu_{ref}$ are reference values, e.g., $\nu_{ref} = 50 \frac{km}{h}$ and $a_{ref} = 1.5 \frac{m}{s^2}$ for urban traffic.
Both reference values can be adjusted with respect to the scenes they evaluated, e.g., traffic-calmed zones
Typical situations with high microscopic scores are fast moving vehicles or situations where the ego vehicle accelerates or brakes.

\subsection{Final Score}
\label{sec:end_score}
The last step is calculating the final score from the four different sub-metrics.
As mentioned in \cite{schutt2022fingerprint}, we use a spacial-related approach instead of finding weights for each term \cite{hallerbach_simulation-based_2018}.
Therefore, the $l_2$-Norm of all four sub-metrics is used:
\begin{equation}
 \mathrm{TQ}_{co} = \sqrt{\mathrm{TQ}_{\Omega}^2 + \mathrm{TQ}_{\eta}^2 + \mathrm{TQ}_{\theta}^2 + \mathrm{TQ}_{\mu}^2}
\end{equation}
However, $\mathrm{TQ}_{co}$ is very sensitive and often marks uncritical situations as critical.
In order to avoid this problem, a penalty term $\rho_{x}$ is introduced:
\begin{equation}
 \mathrm{TQ}_{\rho_{x}} = \rho_{x}\mathrm{TQ}_{co}
\end{equation}
where $\rho_{x}$ can be replaced by a penalty or reward depending on the metric's goal.
In this work, we propose three possible penalty terms with different focus aspects. 
All three terms are based on the distance between the ego vehicle and the nearest vehicle within a scene and punish empty space around an ego vehicle:
\begin{equation}
 \rho_1= \frac{1.5}{d_{min}}
\end{equation}
\begin{equation}
 \rho_2= e^{-\frac{d_{min}}{5.0}}
\end{equation}
\begin{equation}
 \rho_3= e^{-\frac{d_{min}-1.0}{10.0}}
\end{equation}
where $d_{min}$ is the distance to the closest vehicle, respectively.
The proposed criticality threshold for $TQ_{co}$ is $1.5$ and for three combinations with the penalty term $1.0$, where critical scenes achieve a value above the threshold.

$\rho_1$ strongly focuses on distance, and when the ego vehicle gets close to another vehicle, the distance has more influence on the final score than the ego velocity.
However, two close-standing or slow moving vehicles can be seen as less critical than when the ego vehicle is further away but moving fast toward a potential crash.
Therefore, $\rho_2$ and $\rho_3$ are introduced, where $\rho_3$ is less sensitive to the ego passing by standing vehicles.
These penalty terms are situation and map independent. 
Nevertheless, it would be possible to train it regarding the map or situation to get better results if labeled data is available.
This work aims to be as universal as possible and, therefore, this approach was not considered.
\section{Evaluation}
\label{sec:evaluation}

The best way of evaluation would be to compare a new metric with a labeled (urban) traffic dataset where the ground truth of all participants is available regarding their current position/movement and criticality situation.
However, this approach comes with a set of problems.
To our knowledge, there is no available urban traffic dataset with labels regarding the (objective) criticality of single vehicle positions or scenes.
Additionally, large datasets, e.g., INTERACTION dataset \cite{zhan_interaction_2019}, only contain a handful of near-collision and collision situations.
The second problem is that apart from collisions and near-collision situations, it is hard to define an objective and universal set of rules for critical traffic situations.
The wanted criticality often depends on the question at hand or special features within recorded data, e.g., road maps, country-dependent traffic rules, or driver.
Therefore, the task for the evaluation is defined as follows: 

\begin{task}
Critical scenes that can be used for resimulation within the INTERACTION dataset shall be found.
\end{task}

This task is used to derive the following rules regarding the labeling of critical scenes:
\begin{itemize}
    \item[(i)] A collision or near-collision occurred.
    \item[(ii)] Situations have a high crash potential, i.e., minor alterations in velocity or trajectory can lead to collisions in resimulation (e.g., moving cars pass by with a distance of less than $0.5$ meters).
\end{itemize}

The first part of this evaluation in \Cref{sec:data_eval} compares the results of \acl{IUTQ} to a human expert labeled sub-set of the INTERACTION dataset.
The labeled dataset consists of 29.569 scenes, of which 4.263 (14.42\%) are labeled critical.
The \acl{IUTQ}'s results are then compared to other established criticality metrics.
After that, the typical situations found as critical by the \acl{IUTQ} are summarized, and the behavior of the \acl{IUTQ} metric in two known collision and near-collision situations from the INTERACTION dataset is shown in \Cref{sec:eval_typical}.

\subsection{Data-Centered Evaluation}
\label{sec:data_eval}
\begin{table*}[]
\label{tab:evaluation}
\centering
\caption{Comparison of results for selected metrics, showing best entries in bold and worst underlined. Abbreviations are explained in \Cref{sec:data_eval}.}
\begin{tabular}{@{}llllllllllll@{}}
\toprule
    & Dist  & ET                           & GT                           & PET   & PTTC                         & TTC                          & WTTC                         & TQ$\rho_1$    & TQ$\rho_2$           & TQ$\rho_3$ & TQ$_{co}$   \\ \midrule
TP  & 1162  & 3700                         & 745                          & 2374  & 975                          & \underline{605}              & \textbf{4029}                & 3083  & 2149                         & 3530  & 3041  \\
TN  & 22364 & \underline{6212}             & 21592                        & 16261 & \textbf{22366}               & 23626                        & 7373                         & 16627 & 21475                        & 13130 & 13494  \\
FP  & 2942  & 19094                        & 3714                         & 9045  & 2940                         & \textbf{1680}                & \underline{17933}            & 8679  & 3831                         & 12176 & 11812  \\
FN  & 3101  & 563                          & \underline{3518}             & 1889  & 3288                         & 3658                         & \textbf{234}                 & 1180  & 2114                         & 733   & 1222  \\
ACC & 0.796 & \underline{0.335}            & 0.755                        & 0.630 & 0.789                        & \textbf{0.819}               & 0.386                        & 0.667 & 0.799                        & 0.563 & 0.559  \\
MR  & 0.204 & \underline{0.665}            & 0.245                        & 0.370 & 0.211                        & \textbf{0.181}               & 0.614                        & 0.333 & 0.201                        & 0.437 & 0.441  \\
TPR & 0.273 & 0.868                        & 0.175                        & 0.557 & 0.229                        & \underline{0.142}            & \textbf{0.945}               & 0.723 & 0.504                        & 0.828 & 0.713  \\
FPR & 0.116 & 0.755                        & 0.147                        & 0.357 & 0.116                        & \textbf{0.066}               & \underline{0.709}            & 0.343 & 0.151                        & 0.481 & 0.467  \\
TNR & 0.884 & \underline{0.245}            & 0.853                        & 0.643 &  0.884                       & \textbf{0.934}               & 0.291                        & 0.657 & 0.849                        & 0.519 & 0.533  \\
FNR & 0.727 & 0.132                        & 0.825                        & 0.443 & 0.771                        & \underline{0.858}            & \textbf{0.055}               & 0.277 & 0.496                        & 0.172 & 0.287  \\
PRE & 0.283 & \underline{0.162}            & 0.167                        & 0.208 & 0.249                        & 0.265                        & 0.183                        & 0.262 & \textbf{0.359}               & 0.225 & 0.205  \\
CoK & 0.159 & 0.040                        & \underline{0.027}            & 0.117 & 0.116                        & 0.094                        & 0.086                        & 0.220 & \textbf{0.302}               & 0.164 & 0.121  \\
F1S & 0.278 & 0.273                        & \underline{0.171 }           & 0.303 & 0.238                        & 0.185                        & 0.307                        & 0.385 & \textbf{0.419}               & 0.354 & 0.318  \\
MCC & 0.579 & 0.547                        & \underline{0.514}            & 0.572 & 0.558                        & 0.550                        & 0.595                        & 0.636 & \textbf{0.654}               & 0.622 & 0.587  \\\bottomrule
\end{tabular}
\end{table*}

The performance of \acl{IUTQ} is compared to that of other established metrics, such as \ac{TTC} \cite{hayward_near_1972}, \ac{PTTC} \cite{westhofen_criticality_2022}, \ac{WTTC} \cite{wachenfeld_worst-time--collision_2016}, \ac{PET} \cite{Allen.1978}, \ac{GT} \cite{Allen.1978}, \ac{ET} \cite{Allen.1978}, and Euclidean distance (Dist).
For each of these metrics, established thresholds were used, resulting in a binary classification of critical and non-critical situations: $1.5s$ for time-based metrics (except \ac{WTTC}, where $0.47s$ was used), and $1.0m$ for Euclidean distance.

\Cref{tab:evaluation} presents the results for various performance measures, including true positives (TP), true negatives (TN), false negatives (FN), false positives (FP), accuracy (ACC), miss-classification rate (MR), true positive rate (TPR), false positive rate (FPR), true negative rate (TNR), false negative rate (FNR), precision (PRE), \ac{CoK} \cite{wang2019simplified}, \ac{F1S} \cite{chicco2020advantages}, and \ac{MCC} \cite{chicco2020advantages}.
The best performance in each category is indicated by bold entries, while the worst are underlined.
However, it should be noted that only 14.42\% of the imbalanced dataset is labeled as critical. 
Therefore, some bold values may not necessarily indicate good performance of the criticality metric. 
For instance, \ac{WTTC} simply classifies most parts of the data as critical.
\ac{TTC} is another well-performing metric, but its good results are due to the opposite problem: most of the data is classified as uncritical, and with more than 85\% being truly uncritical, some performance measures (e.g., accuracy and misclassification rate) indicate good performance. 
Therefore, we also considered precision, Cohen’s kappa, $F_1$-score, and Matthews correlation coefficient, where higher values indicate better performance. 
In all four performance metrics, the \acl{IUTQ} metric performed best.

Cohen's kappa is a statistical measure that assesses the level of agreement between two raters or classifiers, such as the ground truth and the criticality metric being evaluated in this case \cite{wang2019simplified}.
Its value ranges from $-1$ to $1$, where $0$ indicates that both raters agree by chance, and negative or positive values indicate less or more than chance agreement, respectively. 
It is calculated as follows \cite{wang2019simplified}:
\begin{equation}
\label{equ:kappa}
    \kappa = \frac{p_o - p_e}{1 - p_e} 
\end{equation}
where $P_o$ is the observed relative agreement (accuracy), $p_e$ the hypothetical probability of agreement by chance with data labels randomly assigned.
According to Wang \textit{et al.} \cite{wang2019simplified}, a value of $0.359$, which is the best performance among all the metrics evaluated, is categorized as fair agreement. 
However, the \acl{IUTQ} metric offers an acceptable trade-off between correctly classifying true critical scenarios as critical without including too many uncritical ones in the set of critical scenarios.

Another evaluation metric is the $F_1$-score \cite{chicco2020advantages}, which is a value between $0$ and $1$ that indicates the quality of a binary classification result.
It is calculated as follows \cite{chicco2020advantages}:
\begin{dmath}
    \label{equ:f1score}
    F_1 = \frac{2}{recall^{-1}+precision^{-1}} =
     \frac{2TP}{2TP+FP+FN},
    \end{dmath}
where $TP$  are the true positives, $FP$ false positives, and $FN$ false negatives.
However, the $F_1$-score has limitations in dealing with class imbalance.

To address this problem, Chicco \textit{et al.} \cite{chicco2020advantages} proposed the Matthews correlation coefficient, whose normalized form is calculated as follows:
\begin{dmath}
    \label{equ:nmcc}
    MCC = \frac{(ab - cd)}{\sqrt{(a+c)(a+d)(b+c)(b+d)}},\\
\end{dmath}
where $a=TP$  are the true positives, $b=TN$ true negatives, $c=FP$ false positives, and $d=FN$ false negatives.
the normalized version was used, where the result lies between $0$ and $1$, with $0$ as the worst value, $1$ the best, and $0.5$ equals to a random class assignment.
It should be noted that the Matthews correlation coefficient may decrease when only one class is recognized, as is the case with \ac{TTC} or \ac{WTTC}.

The main reason for the found false positive hits, especially for \acl{IUTQ} with $\rho_2$, lies in situations where vehicles drive towards each other or next to each other with proximity at high velocity even though they are not in the same lane.
The false negatives come from the problem that sometimes the metric is less sensitive to criticality than the labeled data, i.e., the metric recognizes a situation as critical a few timestamps later and categorizes them as uncritical earlier than the labeled data.
This problem in sensitivity arises from the trade-off between too many false positives vs. false negatives, and it depends on the task at hand if the threshold has to be higher to get fewer false positives or lower to get fewer false negatives.
Another way of changing the outcome is to use different penalty terms, where $\rho_1$ emphasizes the criticality close to the vehicle and neglects things that happen further away, $\rho_3$ which takes in more of its surrounding vehicles, and $\rho_2$ as a trade-off in between both options.

It has to be kept in mind, that each metric has its special situations to recognize, and it depends on the question regarding what criticality is measured and which one shall be used.
The main advantage of the \acl{IUTQ} is that it does not specialize in specific vehicle constellations, such as \ac{TTC} (car-following) or \ac{PET} (intersection).
Hence, it can be used in recorded data or simulation without prior knowledge of the scenes or scenarios.

\subsection{Typical Detected Situations and Examples}
\label{sec:eval_typical}

\begin{figure}[t!]
    \centering
     \def\svgwidth{0.99\linewidth}
\begingroup%
  \makeatletter%
  \providecommand\color[2][]{%
    \errmessage{(Inkscape) Color is used for the text in Inkscape, but the package 'color.sty' is not loaded}%
    \renewcommand\color[2][]{}%
  }%
  \providecommand\transparent[1]{%
    \errmessage{(Inkscape) Transparency is used (non-zero) for the text in Inkscape, but the package 'transparent.sty' is not loaded}%
    \renewcommand\transparent[1]{}%
  }%
  \providecommand\rotatebox[2]{#2}%
  \newcommand*\fsize{\dimexpr\f@size pt\relax}%
  \newcommand*\lineheight[1]{\fontsize{\fsize}{#1\fsize}\selectfont}%
  \ifx\svgwidth\undefined%
    \setlength{\unitlength}{293.17736768bp}%
    \ifx\svgscale\undefined%
      \relax%
    \else%
      \setlength{\unitlength}{\unitlength * \real{\svgscale}}%
    \fi%
  \else%
    \setlength{\unitlength}{\svgwidth}%
  \fi%
  \global\let\svgwidth\undefined%
  \global\let\svgscale\undefined%
  \makeatother%
  \begin{picture}(1,0.61815473)%
    \lineheight{1}%
    \setlength\tabcolsep{0pt}%
    \put(0,0){\includegraphics[width=\unitlength,page=1]{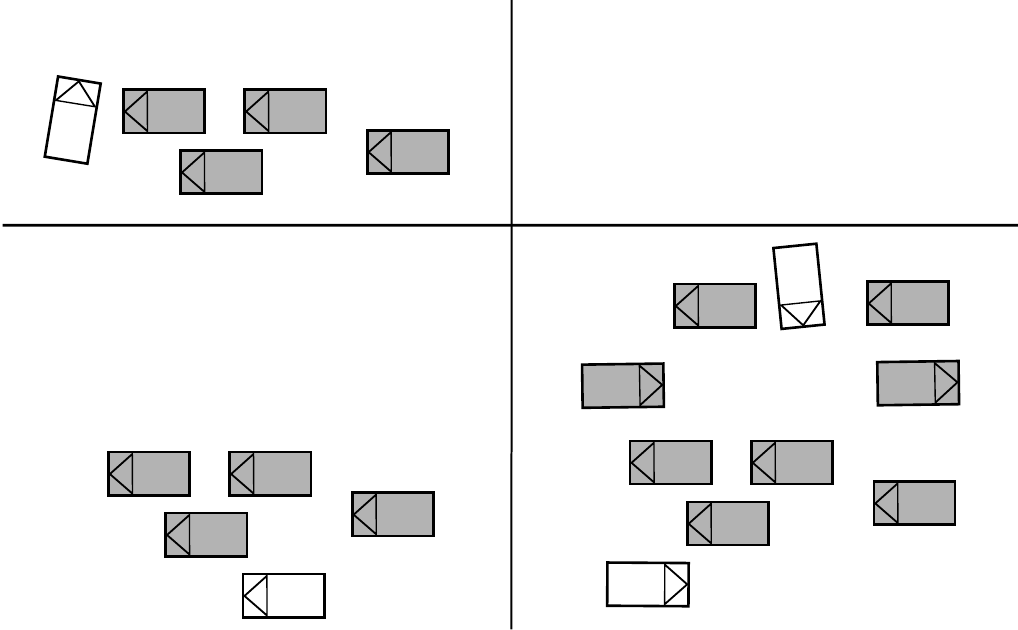}}%
    \put(0.01578695,0.5580318){\color[rgb]{0,0,0}\makebox(0,0)[lt]{\lineheight{1.25}\smash{\begin{tabular}[t]{l}\normalsize{a)}\end{tabular}}}}%
    \put(0.52135068,0.5580318){\color[rgb]{0,0,0}\makebox(0,0)[lt]{\lineheight{1.25}\smash{\begin{tabular}[t]{l}\normalsize{b)}\end{tabular}}}}%
    \put(0,0){\includegraphics[width=\unitlength,page=2]{typical_situations_large.pdf}}%
    \put(0.01693952,0.35285828){\color[rgb]{0,0,0}\makebox(0,0)[lt]{\lineheight{1.25}\smash{\begin{tabular}[t]{l}\normalsize{c)}\end{tabular}}}}%
    \put(0.52135075,0.35285828){\color[rgb]{0,0,0}\makebox(0,0)[lt]{\lineheight{1.25}\smash{\begin{tabular}[t]{l}\normalsize{d)}\end{tabular}}}}%
    \put(0.01540732,0.14768476){\color[rgb]{0,0,0}\makebox(0,0)[lt]{\lineheight{1.25}\smash{\begin{tabular}[t]{l}\normalsize{e)}\end{tabular}}}}%
    \put(0.52555765,0.14768476){\color[rgb]{0,0,0}\makebox(0,0)[lt]{\lineheight{1.25}\smash{\begin{tabular}[t]{l}\normalsize{f)}\end{tabular}}}}%
  \end{picture}%
\endgroup%

\caption{Typical scenes classified as critical. The faster the ego vehicle (white) or the higher the difference in velocity among all traffic participants, the higher the criticality.}
\label{fig:situations_typical}
\end{figure}

\begin{figure*}[t!]
    \centering
       \def\svgwidth{0.99\textwidth}
       {\footnotesize
    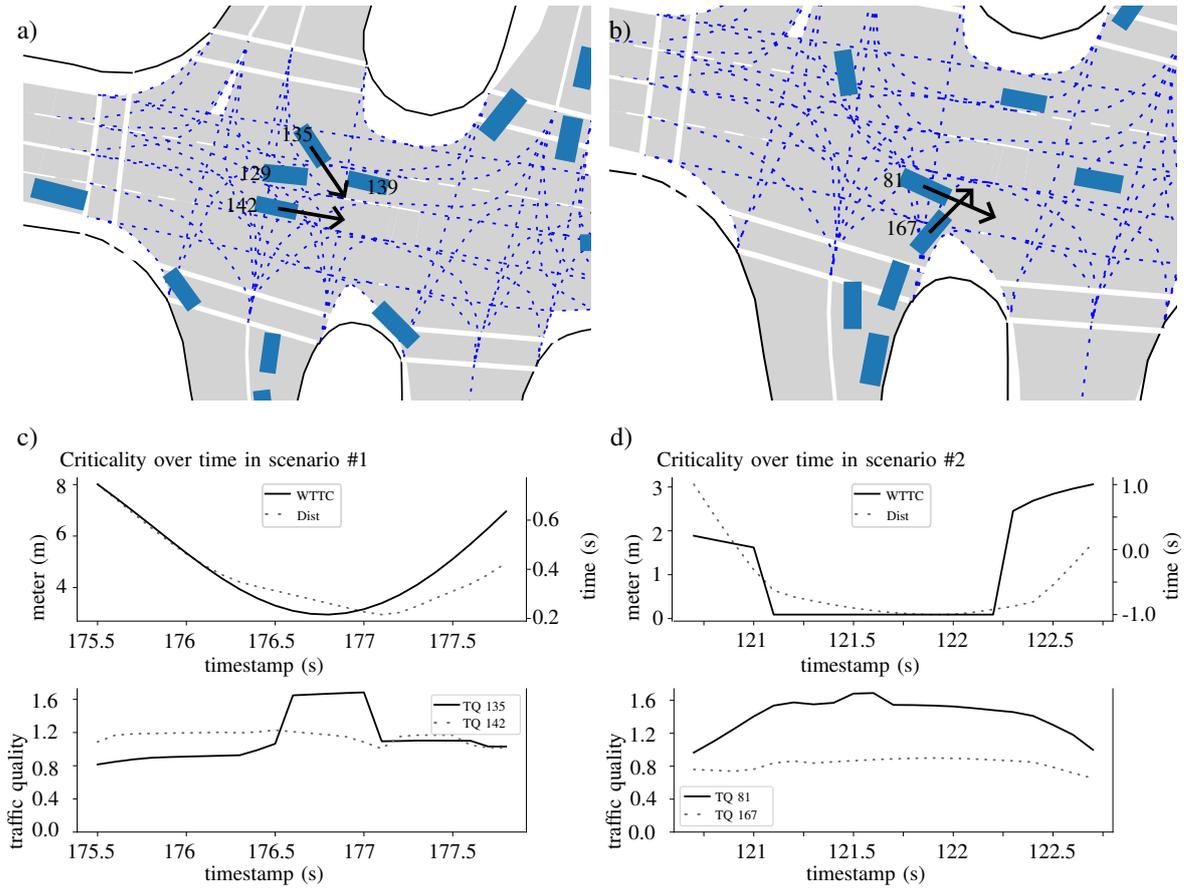}
\caption{(a) shows a snapshot of a near-collision situation (scenario \#1). (b) shows a snapshot of a collision with two intersection trajectories (scenario \#2). (c) and (d) show the progression of criticality metrics (distance, \ac{WTTC}, traffic quality) over time, respectively.}
\label{fig:example_scenarios}
\vspace{-1ex}
\end{figure*}


\subsubsection{Typical Critical Situations}
Upon examining the traffic in the INTERACTION dataset, typical situations have been detected where IUTQ indicates criticality.
In general, stationary objects are considered less critical than moving objects since they are assumed to have less influence on the outcome of a situation than moving objects. 
Often, these situations exhibit \textit{high-tension}, characterized by dense traffic with varying velocities of different traffic participants, as shown in  \Cref{fig:situations_typical}.
These \textit{high-tension} situations typically include:
\begin{itemize}
    \item[(i)] The ego vehicle passing by or driving around one or more stationary or slow-moving objects:
    \begin{itemize}
        \item Other vehicles waiting to take a left or right turn  (see \Cref{fig:situations_typical} (a)).
        \item The ego vehicle approaching the end of already waiting vehicles (see \Cref{fig:situations_typical} (b)) or accelerating at the start of a traffic jam (see \Cref{fig:situations_typical} (c)).
        \item The ego vehicle moving around parked or slow cars  (see \Cref{fig:situations_typical} (d)).
    \end{itemize}
    The IUTQ value increases with proximity, ego velocity, and velocity differences between vehicles.
    \item[(ii)] Vehicles moving towards or next to each other in the same lane (see \Cref{fig:situations_typical} (e)) or oncoming directions (see \Cref{fig:situations_typical} (f)). This scenario also includes cases where both vehicles are on different lanes. Since no map data is used due to the simplified calculation, it is impossible to distinguish if both vehicles are in the same lane.
\end{itemize}

\subsubsection{Near-collision Example}

In the near-collision scenario depicted in\Cref{fig:example_scenarios} (a), two vehicles (129, 139) are stationary at an intersection.
While vehicle 135 attempts to cross the intersection between both mentioned vehicles, vehicle 142 passes 129 and 139 in an oncoming lane. 
All of this occurs in less than three seconds, and 142 cannot see 135 early on. 
Both vehicles barely avoid a collision.
\Cref{fig:example_scenarios} (c) presents the measured distance and \ac{WTTC} between 135 and 142, respectively.
The minimum distance between the two is approximately $2.75m$, which, in slow urban traffic, does not indicate a critical situation.
However, their WTTC drops to $0.22s$, representing a critical situation.
The \acl{IUTQ} value for vehicle 135 begins to increase when it passes through the gap between 129 and 139 and remains slightly above $1.0$ throughout the maneuver.
The \acl{IUTQ} value for 142 stays around $1.2$, with a slight dip when vehicle 135 comes to a halt and accelerates again.

\subsubsection{Collision Example}
The second example, illustrated in \Cref{fig:example_scenarios} (b), is taken from the same dataset and involves vehicle 81 coming from the left side of an intersection and turning right onto the bottom arm, while vehicle 167 from the left bottom arm intends to turn right as well.
In this case, vehicle 81 collides with vehicle 164, and both vehicles pull over to the side of the road, which is a common post-collision behavior. 
\Cref{fig:example_scenarios} (d) shows the distance curves between vehicle 81 and 167, which decrease to $0m$ and result in a \ac{WTTC} value of $0s$ at timestamp $121.1$.
The negative \ac{WTTC} value indicates that after the crash, both vehicles came to a stop, and the system of equations used to calculate the \ac{WTTC} cannot be solved.
The \acl{IUTQ} of vehicle 81 is constantly high due to the critical maneuver and collision.
The IUTQ of vehicle 81 remains constantly high throughout the critical maneuver and collision, whereas the value of vehicle 167 remains below $1.0$ as it is still stationary and, therefore, has a lower impact on the outcome of the situation.

\subsection{Limitations}
The penalty term plays an crucial role in the overall result.
The traffic quality neglects critical situations near the ego vehicle in favor of dense traffic elsewhere on the map if there is no penalty term.
The penalty term $\rho_2$ favors critical situations within a five-meter radius around the ego vehicle.
It is possible to find a better radius, particularly if it is optimized by a machine-learning algorithm that uses labeled data for training.
However, this approach may result in overfitting to a certain map variation, e.g., intersections.
Therefore, the proposed penalty term is a fair trade-off for the desired goal of situation independence.

Situations like those shown in \Cref{fig:situations_typical} (c) may not be considered critical since the ego vehicle is at the beginning of a line of standing cars.
However, the metric is unable to distinguish between situations like those in \Cref{fig:situations_typical} (b) and \Cref{fig:situations_typical} (c).

The metric was not tested with vulnerable road users, such as bicycles or pedestrians, as the INTERACTION dataset does not contain such data.
Consequently, objectively critical situations involving vulnerable road users may be overlooked, e.g., a collision with a pedestrian might not be recognized.

Furthermore, our evaluation only utilized urban traffic data, and the performance of the metric was not tested with highway scenarios and situations. 
For highway data, we recommend referring to the work of Hallerbach \textit{et al.} \cite{hallerbach_simulation-based_2018}, as they have proposed a similar metric for highway scenarios.

\section{Conclusion}
\label{sec:conclusion}
The fundamental requirement for a metric to assess the criticality of a complex and congested traffic scene is to be agnostic of vehicle constellations and a predefined set of adversary vehicles. 
This approach has been explained, discussed, and evaluated in this work.
The presented metric does not necessitate knowledge of which vehicle is considered concerning the ego vehicle, as all traffic objects are accounted for in the metric.  
Thus, it represents a holistic approach to scene evaluation.
The \acl{IUTQ}  metric is suitable for assessing and identifying critical and high-tension scenes in large real-world datasets or scenario exploration.
Unlike other established metrics, it does not compute the metric between the ego and every other adversary individually.
It uses a penalty term that can be adjusted according to the needs of the user.
If no adjustment is needed we propose $\rho_2$ as the default penalty term.
In case labeled data is available and the map is not chaining, we suggest to use a trained weights approach for each sub-metric as proposed by Hallerbach \textit{et al.} \cite{hallerbach_simulation-based_2018}.
It is generally challenging for a single metric to objectively identify all critical situations.
As we proposed in prior work \cite{schutt2022fingerprint}, we advocate for a combined approach.
Nevertheless, a universal metric like the \acl{IUTQ} serves as a foundation for evaluating a dataset without prior knowledge of scene criticality. 
Additionally, this metric can be employed for scenario exploration and generation where several actors are involved in increasing the criticality.


\section*{Acknowledgment}
The research leading to these results is funded by the German Federal Ministry for Economic Affairs and Climate Action within the projects \textit{Verifikations- und Validierungsmethoden automatisierter Fahrzeuge im urbanen Umfeld} a project from the PEGASUS family, based on a decision by the Parliament of the Federal Republic of Germany. The authors would like to thank the consortium for the successful cooperation.

\bibliographystyle{IEEEtran}
\bibliography{00_main}

\begin{thebibliography}{10}
\providecommand{\url}[1]{#1}
\csname url@samestyle\endcsname
\providecommand{\newblock}{\relax}
\providecommand{\bibinfo}[2]{#2}
\providecommand{\BIBentrySTDinterwordspacing}{\spaceskip=0pt\relax}
\providecommand{\BIBentryALTinterwordstretchfactor}{4}
\providecommand{\BIBentryALTinterwordspacing}{\spaceskip=\fontdimen2\font plus
\BIBentryALTinterwordstretchfactor\fontdimen3\font minus
  \fontdimen4\font\relax}
\providecommand{\BIBforeignlanguage}[2]{{%
\expandafter\ifx\csname l@#1\endcsname\relax
\typeout{** WARNING: IEEEtran.bst: No hyphenation pattern has been}%
\typeout{** loaded for the language `#1'. Using the pattern for}%
\typeout{** the default language instead.}%
\else
\language=\csname l@#1\endcsname
\fi
#2}}
\providecommand{\BIBdecl}{\relax}
\BIBdecl

\bibitem{EUtypeapprovannex}
{Council of European Union}, ``{Commission Implementing Regulation (EU)
  2022/1426 of 5 August 2022},''
  \url{https://eur-lex.europa.eu/legal-content/EN/TXT/HTML/?uri=CELEX:32022R1426#d1e41-20-1},
  2022, [Online; accessed 17-October-2022].

\bibitem{EUtypeapprov}
------, ``{Regulation (EU) 2019/2144 of the European Parliament and of the
  Council of 27 November 2019 },''
  \url{https://eur-lex.europa.eu/eli/reg/2019/2144/oj}, 2019, [Online; accessed
  17-October-2022].

\bibitem{schutt2022fingerprint}
B.~Sch{\"u}tt, M.~Zipfl, J.~M. Z{\"o}llner, and E.~Sax, ``Fingerprint of a
  traffic scene: an approach for a generic and independent scene assessment,''
  in \emph{2022 International Conference on Electrical, Computer,
  Communications and Mechatronics Engineering (ICECCME)}.\hskip 1em plus 0.5em
  minus 0.4em\relax IEEE, 2022, pp. 1--8.

\bibitem{ulbrich_defining_2015}
\BIBentryALTinterwordspacing
S.~Ulbrich, T.~Menzel, A.~Reschka, F.~Schuldt, and M.~Maurer,
  ``\BIBforeignlanguage{en}{Defining and {Substantiating} the {Terms} {Scene},
  {Situation}, and {Scenario} for {Automated} {Driving}},'' in
  \emph{\BIBforeignlanguage{en}{2015 {IEEE} 18th {International} {Conference}
  on {Intelligent} {Transportation} {Systems}}}.\hskip 1em plus 0.5em minus
  0.4em\relax Gran Canaria, Spain: IEEE, Sep. 2015, pp. 982--988. [Online].
  Available: \url{http://ieeexplore.ieee.org/document/7313256/}
\BIBentrySTDinterwordspacing

\bibitem{schutt2022taxonomy}
B.~Sch{\"u}tt, M.~Steimle, B.~Kramer, D.~Behnecke, and E.~Sax, ``A taxonomy for
  quality in simulation-based development and testing of automated driving
  systems,'' \emph{IEEE Access}, vol.~10, pp. 18\,631--18\,644, 2022.

\bibitem{neurohr2021criticality}
C.~Neurohr, L.~Westhofen, M.~Butz, M.~H. Bollmann, U.~Eberle, and R.~Galbas,
  ``Criticality analysis for the verification and validation of automated
  vehicles,'' \emph{IEEE Access}, vol.~9, pp. 18\,016--18\,041, 2021.

\bibitem{junietz2019microscopic}
P.~M. Junietz, ``Microscopic and macroscopic risk metrics for the safety
  validation of automated driving,'' Ph.D. dissertation, Technische
  Universit{\"a}t Darmstadt, 2019.

\bibitem{westhofen_criticality_2022}
\BIBentryALTinterwordspacing
L.~Westhofen, C.~Neurohr, T.~Koopmann, M.~Butz, B.~Schütt, F.~Utesch,
  B.~Neurohr, C.~Gutenkunst, and E.~Böde, ``Criticality {Metrics} for
  {Automated} {Driving}: {A} {Review} and {Suitability} {Analysis} of the
  {State} of the {Art},'' Springer Nature, Tech. Rep., Jun. 2022. [Online].
  Available: \url{https://doi.org/10.1007/s11831-022-09788-7}
\BIBentrySTDinterwordspacing

\bibitem{hallerbach_simulation-based_2018}
S.~Hallerbach, Y.~Xia, U.~Eberle, and F.~Koester,
  ``\BIBforeignlanguage{en}{Simulation-{Based} {Testing} of {Cooperative} and
  {Automated} {Vehicles}},'' \emph{\BIBforeignlanguage{en}{SAE International
  Journal of Connected and Automated Vehicles}}, vol.~1, no. 2018-01-1066, pp.
  93--106, 2018.

\bibitem{zhan_interaction_2019}
\BIBentryALTinterwordspacing
W.~Zhan, L.~Sun, D.~Wang, H.~Shi, A.~Clausse, M.~Naumann, J.~Kummerle,
  H.~Konigshof, C.~Stiller, A.~de~La~Fortelle, and M.~Tomizuka,
  ``\BIBforeignlanguage{en}{{INTERACTION} {Dataset}: {An} {INTERnational},
  {Adversarial} and {Cooperative} {moTION} {Dataset} in {Interactive} {Driving}
  {Scenarios} with {Semantic} {Maps}},''
  \emph{\BIBforeignlanguage{en}{arXiv:1910.03088 [cs, eess]}}, Sep. 2019,
  arXiv: 1910.03088. [Online]. Available: \url{http://arxiv.org/abs/1910.03088}
\BIBentrySTDinterwordspacing

\bibitem{hayward_near_1972}
J.~C. Hayward, ``Near miss determination through use of a scale of danger,'' in
  \emph{Proc. 51st Annu. Meeting Highway Res. Board}, 1972.

\bibitem{wachenfeld_worst-time--collision_2016}
\BIBentryALTinterwordspacing
W.~Wachenfeld, P.~Junietz, R.~Wenzel, and H.~Winner,
  ``\BIBforeignlanguage{en}{The worst-time-to-collision metric for situation
  identification},'' in \emph{\BIBforeignlanguage{en}{2016 {IEEE} {Intelligent}
  {Vehicles} {Symposium} ({IV})}}.\hskip 1em plus 0.5em minus 0.4em\relax
  Gotenburg, Sweden: IEEE, Jun. 2016, pp. 729--734. [Online]. Available:
  \url{http://ieeexplore.ieee.org/document/7535468/}
\BIBentrySTDinterwordspacing

\bibitem{Allen.1978}
B.~L. Allen, B.~T. Shin, and P.~J. Cooper, ``{Analysis of Traffic Conflicts and
  Collisions},'' \emph{{Transp. Res. Rec.}}, vol. 667, pp. 67--74, 1978.

\bibitem{wang2019simplified}
J.~Wang, Y.~Yang, and B.~Xia, ``A simplified cohen’s kappa for use in binary
  classification data annotation tasks,'' \emph{IEEE Access}, vol.~7, pp.
  164\,386--164\,397, 2019.

\bibitem{chicco2020advantages}
D.~Chicco and G.~Jurman, ``The advantages of the matthews correlation
  coefficient (mcc) over f1 score and accuracy in binary classification
  evaluation,'' \emph{BMC genomics}, vol.~21, no.~1, pp. 1--13, 2020.

\end{thebibliography}
%

\end{document}